\documentclass{article} 
\usepackage{iclr2025_conference,times}

\usepackage{amsmath,amsfonts,bm}

\def\eqref#1{equation~\ref{#1}}

\def\1{\bm{1}}

\DeclareMathAlphabet{\mathsfit}{\encodingdefault}{\sfdefault}{m}{sl}
\SetMathAlphabet{\mathsfit}{bold}{\encodingdefault}{\sfdefault}{bx}{n}

\usepackage{hyperref}
\usepackage{url}
\usepackage{microtype}

\usepackage{graphicx}
\usepackage{subcaption}
\usepackage{booktabs} 
\usepackage{csquotes}
\usepackage{longtable}

\usepackage{amsmath}
\usepackage{amssymb}
\usepackage{mathtools}
\usepackage{amsthm}
\usepackage{xspace}
\usepackage{colortbl}
\usepackage{multirow}
\usepackage{wrapfig}
\usepackage{float}
\usepackage{placeins}
\usepackage{algpseudocode}
\usepackage{algorithm}
\usepackage{multirow}
\usepackage{lineno}
\usepackage{stackengine}
\usepackage{caption}
\usepackage{rotating}
\usepackage{wrapfig}
\usepackage[labelfont={bf}]{caption}

\usepackage{enumitem}
\setlist[itemize]{noitemsep}

\newcommand{\mname}{GraphPINE}
\newcommand{\fullname}{\textbf{Graph} \textbf{P}ropagating \textbf{I}mportance \textbf{N}etwork for \textbf{E}xplanation}
\newcommand{\repo}{https://anonymous.4open.science/r/GraphPINE-40DE}

\title{
    \mname: Graph importance propagation for interpretable drug response prediction 
}

\author{Yoshitaka Inoue \\
Department of Computer Science and Engineering\\
University of Minnesota\\
Minneapolis, MN, USA \\
Computational Biology Branch, National Library of Medicine\\
Developmental Therapeutics Branch, National Cancer Institute\\
Bethesda, MD, USA \\
\texttt{inoue019@umn.edu} \\
\And
Tianfan Fu \\
Department of Computer Science \\
Nanjing University \\
Nanjing, Jiangsu, China \\
\texttt{futianfan@gmail.com}
\And
Augustin Luna \\
Computational Biology Branch, National Library of Medicine \\
Developmental Therapeutics Branch, National Cancer Institute \\
Bethesda, MD, USA \\
\texttt{augustin@nih.gov}
}

\iclrfinalcopy 
\begin{document}

\maketitle

\begin{abstract}

Explainability is necessary for many tasks in biomedical research. Recent explainability methods have focused on attention, gradient, and Shapley value. These do not handle data with strong associated prior knowledge and fail to constrain explainability results based on known relationships between predictive features.

We propose GraphPINE, a graph neural network (GNN) architecture leveraging domain-specific prior knowledge to initialize node importance optimized during training for drug response prediction. Typically, a manual post-prediction step examines literature (i.e., prior knowledge) to understand returned predictive features. While node importance can be obtained for gradient and attention after prediction, node importance from these methods lacks complementary prior knowledge; GraphPINE seeks to overcome this limitation. GraphPINE differs from other GNN gating methods by utilizing an LSTM-like sequential format. We introduce an importance propagation layer that unifies 1) updates for feature matrix and node importance and 2) uses GNN-based graph propagation of feature values. This initialization and updating mechanism allows for informed feature learning and improved graph representation. 

We apply GraphPINE to cancer drug response prediction using drug screening and gene data collected for over 5,000 gene nodes included in a gene-gene graph with a drug-target interaction (DTI) graph for initial importance. The gene-gene graph and DTIs were obtained from curated sources and weighted by article count discussing relationships between drugs and genes. GraphPINE achieves a PR-AUC of 0.894 and ROC-AUC of 0.796 across 952 drugs. Code is available at~\url{https://anonymous.4open.science/r/GraphPINE-40DE}.
\end{abstract}

\section{Introduction}

Drug response prediction (DRP) is an open research challenge in personalized medicine and drug discovery. Work in this research area seeks to improve treatment outcomes and reduce adverse effects.  However, the complex interplay between drug compounds and cellular entities makes this task challenging. Traditional approaches often fail to capture the intricate network of interactions that influence drug response, leading to suboptimal predictions with limited interpretability. Despite recent advancements, current DRP methods face challenges such as data heterogeneity, limited sample sizes, and the need for multi-omics integration~\citep{azuaje2017computational, lu2018multi, vamathevan2019applications}. 

Greater data availability combined with algorithmic improvements have led to an increase in machine learning (ML) techniques in this research area. GNNs have emerged as a promising approach due to their ability to model complex relational data~\citep{kipf2016semi}. Recent GNN variants, such as Graph Transformer Networks~\citep{yun2019graph} and Graph Diffusion Networks~\citep{klicpera2019diffusion}, have shown promise in capturing complex, long-range dependencies in biological networks. However, these advanced architectures often come at the cost of increased complexity and reduced interpretability. This leads to two main limitations in existing GNN models for DRP. First, many models do not incorporate known biological information, such as DTI. This omission can lead to predictions that, while accurate, may not align with known biological mechanisms. Second, the ``black box'' nature of many deep learning models makes it difficult for researchers and clinicians to understand and trust the predictions. This lack of interpretability is a significant barrier to adopting these models for furthering understanding of drug mechanisms.

While some attempts have been made to incorporate biological priors into GNNs~\citep{zitnik2018modeling} or improve interpretability~\citep{ying2019gnnexplainer}, no existing method addresses both challenges in the context of DRP. To address these limitations, we introduce \mname~(\fullname), a novel GNN approach combining the predictive power of deep learning with biologically informed feature importance propagation and interpretability.

The key innovation of \mname~lies in its Importance Propagation (IP) Layer, which updates and propagates gene importance scores across the network during the learning process. This mechanism allows \mname~to:
\vspace{-5pt}
\begin{enumerate}[leftmargin=*]
    \item Integrate known DTI information with the underlying gene network structure, ensuring the model's predictions are grounded in known biological interactions.
    \item Capture drug-gene interactions with N-hops GNN layers, providing a more comprehensive view of drug influence on the gene network.
    \item Generate interpretable visualizations of gene-gene interactions under the drug treatment, offering new perspectives on potential drug action mechanisms.
\end{enumerate}
\vspace{-10pt}



\section{Related Works}\label{app:related_works}


\subsection{Drug Response Prediction}

DRP refers to the process of forecasting how a particular drug will affect the viability of a biological system based on various data inputs such as genomic information and molecular structures~\citep{adam2020machine}. The goal is to predict the drug sensitivity, which can aid in personalized medicine, allowing for more targeted treatments for patients. 

Several notable models have emerged: \citet{li2019deepdsc} developed DeepDSC combining an autoencoder for gene expression to obtain hidden embeddings, which are then used as input to a feed-forward network along with drug fingerprint embeddings. \citet{lao2024deepaeg} implemented the DeepAEG, including transformer for SMILES and attention for multi-omics data (e.g., mutation, gene expression). 

\subsection{Graph Neural Networks in Computational Biology}

GNNs have emerged as a powerful tool for modeling complex biological systems. \citet{zitnik2018modeling} utilized GNNs for side effect prediction with drug-drug interaction networks. GNNs have also been used for molecular property prediction, showcasing the potential of GNNs in cheminformatics \citep{fu2021mimosa}. For the DRP, GraphDRP~\citep{nguyen2021graph} integrates gene expression and protein-protein interaction networks, while MOFGCN~\citep{peng2021predicting} combines multi-omics data. 

\subsection{Explainable AI in Biological Applications}

As ML models become complex, there is a growing need for interpretability, especially in biomedical applications where understanding the rationale behind predictions is fundamental for clinical research. Explainable AI methods can be categorized into three main types:

\vspace{-5pt}
\begin{enumerate}[leftmargin=*]
    \item \textbf{Gradient-based methods:} These techniques utilize gradient information to highlight important features. For example, Grad-CAM~\citep{selvaraju2020grad} generates visual explanations for decisions made by convolutional neural networks.  \cite{fu2021differentiable} produces molecular substructure-level gradient to provide interpretability for drug design. 
\vspace{-5pt}
    \item \textbf{Attention-based methods:} These approaches leverage attention coefficients to identify relevant inputs. \cite{abnar2020quantifying} propose attention flow to quantify the information propagation through self-attention layers, improving the interpretability of the Transformer. For DRP, \cite{inoue2024drgat} employs Graph Attention Network (GAT)~\citep{velivckovic2017graph} on a heterogeneous network of proteins, cell lines, and drugs, offering interpretability through attention coefficients. 
\vspace{-5pt}

    \item \textbf{Shapley value-based methods:} SHapley Additive exPlanations (SHAP)~\citep{lundberg2017unified} assigns importance values to input features based on game theory principles, providing a unified measure of feature contributions to model predictions.
\end{enumerate}
\vspace{-5pt}

\mname~is related to the attention-based methods but with key distinctions. Unlike typical attention mechanisms assigning importance to edges, \mname~uses DTI information to initialize node importance scores. It propagates this importance throughout the learning process along with the graph structure. This approach incorporates biological knowledge, thereby enhancing interpretability.

\subsection{Information Propagation in Neural Networks}

\cite{Shrikumar2017} proposed DeepLIFT (Deep Learning Important FeaTures). This method computes importance scores, capturing non-linear dependencies that might be missed by other approaches. DeepLIFT addresses the limitations of traditional gradient-based methods by considering the difference from a reference input. This approach offers a more nuanced understanding of feature contributions and provides more interpretable explanations of model outputs.

More recently, \cite{abnar2020quantifying} introduced Attention Flow, a method designed for Transformer models. This approach models the propagation of attention through the layers of a Transformer, quantifying how information flows from input tokens to output tokens. Attention Flow provides a more accurate measure of token relationships compared to raw attention weights, offering insights into how Transformer models process and utilize information across their multiple attention layers.

These methods can all be viewed as specialized forms of information propagation. In each case, the ``information'' being propagated represents the relevance, importance, or attention associated with different components of the network. These approaches demonstrate how the concept of information propagation can be leveraged to enhance the interpretability of complex neural network models, offering valuable insights into their decision-making processes across various network architectures.

\subsection{Importance Gating with GNNs}

Recent studies have proposed different approaches for incorporating gating mechanisms into GNNs. Two notable examples are Event Detection GCN \citep{lai2020event} and CID-GCN \citep{zeng2021cid}.

Event detection is a natural language processing (NLP) task aiming to identify specific events (e.g., accidents) from documents. Event Detection GCN implements a gating mechanism utilizing trigger candidate information (e.g., potential event-indicating words: "attacked") to filter noise from hidden vectors. The model incorporates gate diversity across layers and leverages syntactic importance scores from dependency trees, which represent grammatical relationships between words in sentences. 

CID-GCN, designed for chemical-disease relation extraction, constructs a heterogeneous graph with mentions (representing specific entity occurrences), sentences (containing the textual context), and entities (normalizing multiple mentions) nodes. The model employs gating mechanisms to address the over-smoothing problem and enables effective information propagation between distant nodes.

\mname~advances these concepts through two key ideas. First, it introduces a novel approach to importance scoring by leveraging domain-specific prior knowledge for initialization rather than relying solely on previous hidden states. Second, it implements a unified importance score updating mechanism through graph learning, departing from the context-based or two-step gating methods.

\section{Methods}\label{sec:method}

\begin{figure}[t]
\centering
\includegraphics[width=\textwidth,keepaspectratio]{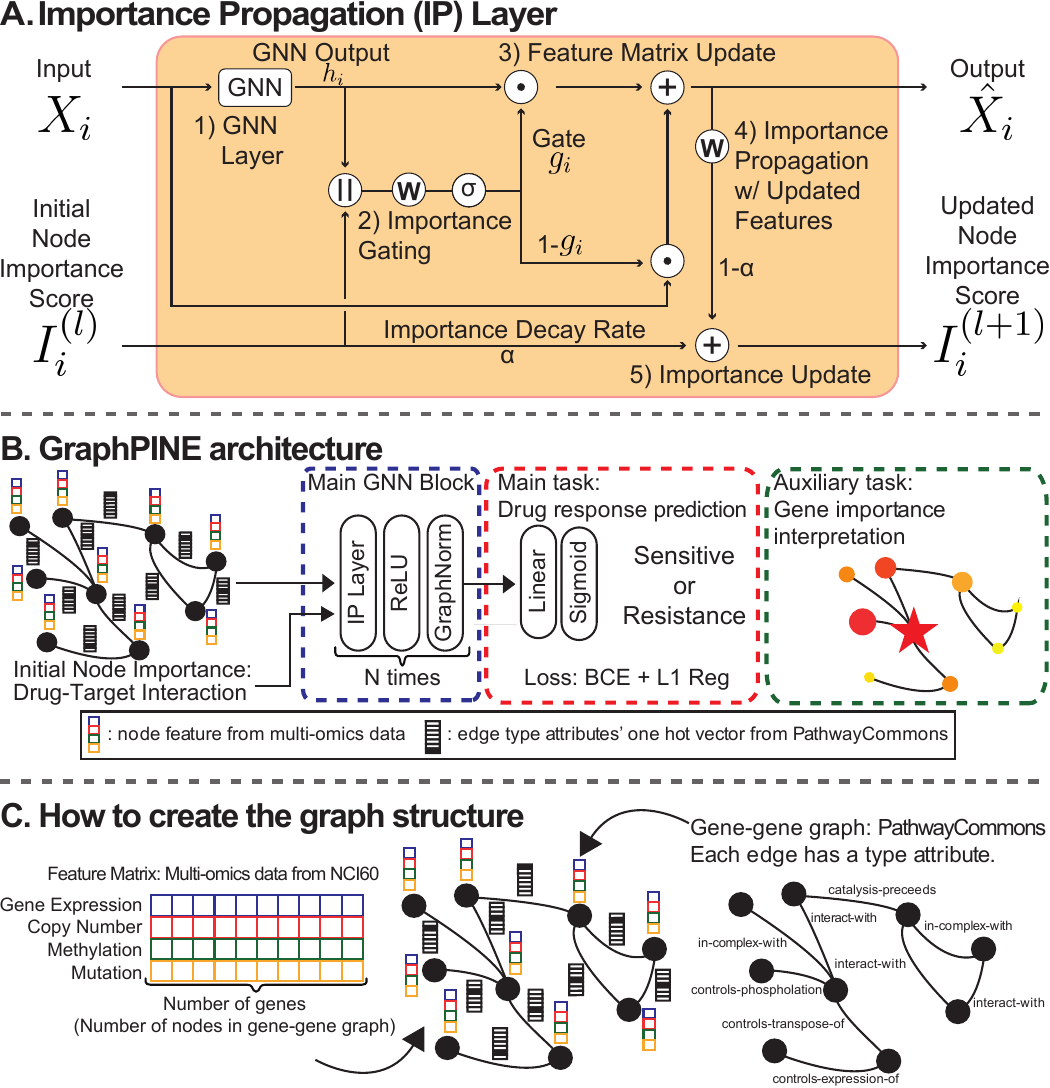}
\caption{\textbf{Overview of \mname~Components.} (A)  Importance Propagation (IP) Layer: This illustrates the key components of the IP Layer in the \mname~model, including the GNN, importance gating, feature updates with residual connections, importance propagation, and updates. The symbols represent the following operations: $\sigma$ is the activation function, $\odot$ is element-wise multiplication, $\times$ is multiplication, $+$ is addition, $W$ denotes weighted calculation with bias, $||$ represents concatenation, and $\alpha$ is a hyperparameter for controlling importance. (B) \mname~architecture. (C) Data Creation Overview: The model integrates multi-omics data (gene expression, copy number, methylation, mutation) from NCI60~\citep{shoemaker2006nci60} with gene-gene interaction networks from PathwayCommons~\citep{cerami2010pathway, rodchenkov2020pathway}. Each edge has attributes such as ``interact-with'', which are converted into one-hot vectors for edge attribution.}
\label{fig:oracle_framework}
\vspace{-20pt}
\end{figure}

This section presents the \mname~model, including data preprocessing, network construction, and the model architecture. \mname~is a GNN architecture designed for accurate and interpretable DRP, leveraging multi-omics data (e.g., gene expression, copy number variation, methylation, and mutation information), along with known biological interactions to provide comprehensive insights into drug-target relationships, as illustrated in Figure~\ref{fig:oracle_framework}.

\subsection{Data Preprocessing and Network Construction}

We integrated three key datasets to generate a gene-gene interaction network with initial importance weights. First, we incorporated a gene-gene interaction network from PathwayCommons as our base graph structure. Second, we collected multi-omics profiles from NCI-60 cell lines to serve as node features. Third, we obtained drug-target interactions from five databases: CTD~\citep{davis2023comparative}, DrugBank~\citep{Wishart2018DrugBank}, DGIdb~\citep{freshour2021integration}, STITCH~\citep{szklarczyk2016stitch}, and KIBA~\citep{Tang2014KIBA}, which we used to establish initial node importance weights. We selected genes based on their variance, network centrality, and drug-target interaction frequency. Comprising 5,181 genes and 630,632 interactions. In addition, network edge types were encoded as one-hot vectors (see Appendix \ref{app:data}).

Gene expression data was normalized using TPM, Log2 transformation, and winsorization. Each gene in each cell line was represented by a 4-dimensional feature vector combining all multi-omics data. DTI scores were calculated from multiple databases, encompassing both direct physical binding between drugs and targets, as well as their indirect associations.

Let $S_{dti}(d_i, g_j)$ be the initial importance score for drug $d_i$ and gene $g_j$. We normalized these scores to a range of [0.5, 1]:

\vspace{-10pt}
\begin{equation}
\begin{aligned}
\text{log\_count} &= \log(1 + \text{PubMed ID\_count}) \\
S_{dti}(d_i, g_j) &= 0.5 + 0.5 \times \frac{\text{log\_count} - \min(\text{log\_count})}{\max(\text{log\_count}) - \min(\text{log\_count})}. 
\end{aligned}
\end{equation}
\vspace{-10pt}

Here, log\_count refers to the log-transformed PubMed ID counts, where PubMed ID\_count represents the number of papers retrieved from PubMed ESearch\citep{sayers2009utilities} using the query that searches for co-mentions by combining drug name and gene name. $d_i$ and $g_j$ denote specific drugs and genes. The 0.5 is added to distinguish the genes that are in databases, but they don't have the literature information. Therefore, the range of $S_{dti}(d_i, g_j)$ is $S_{d t i}\left(d_i, g_j\right) \in\{0\} \cup[0.5,1]$.

\subsection{\mname~Model Architecture}

The \mname~model predicts drug response and learns gene importance using a gene interaction network $G = (V, E)$ with node features $X \in \mathbb{R}^{|V| \times d}$; edge features $E_{attr} \in \mathbb{R}^{|E| \times f}$, and importance scores $I \in \mathbb{R}^{|V|}$. The model outputs a predicted drug response $\hat{y} \in \mathbb{R}$ and updated importance score $I' \in \mathbb{R}^{|V|}$, utilizing edge-aware GNN architectures (i.e., Graph Attention Network (GAT)~\citep{velivckovic2017graph}, Graph Transformer (GT)~\citep{yun2019graph}, and Graph Isomorphism Network with Edge features (GINE)~\citep{hu2019strategies}).

\subsubsection{Importance Propagation Layer}

The Importance Propagation Layer (IP Layer) is a key component that processes and updates node features while considering their importance scores. The layer operates through five main steps:

1. Apply TransformerConv to process node features. This step transforms the input node features using graph topology:
\begin{equation}\label{eq:gat}
\mathbf{h}_i = \text{TransformerConv}(\mathbf{x}_i, \text{edge\_index}, \text{edge\_attr})
\end{equation}

2. Generate gate using GT output and importance scores. The gate controls information flow based on node importance:
\begin{equation}
\mathbf{g}_i = \sigma(\mathbf{W}_g[\mathbf{h}_i \| I_i] + \mathbf{b}_g)
\end{equation}
where $\sigma$ is the sigmoid function and $\|$ denotes concatenation.

3. Update node features using a gating mechanism that combines original and transformed features:
\begin{equation}
\mathbf{\hat{x}}_i = \mathbf{g}_i \odot \mathbf{h}_i + (1 - \mathbf{g}_i) \odot \mathbf{x}_i
\end{equation}
where $\odot$ represents element-wise multiplication.

4. Propagate importance scores through the network using a learnable transformation:
\begin{equation}
I'_i = \mathbf{W}_p\mathbf{\hat{x}}_i + b_p
\end{equation}

5. Update and normalize importance scores. First, update scores using a decay mechanism:
\begin{equation}
I_i^{(l+1)} = \alpha I_i^{(l)} + (1 - \alpha) I_i'^{(l)}
\end{equation}
Then, normalize and threshold the scores:
\begin{equation}\label{eq:propagation}
\begin{aligned}
    I_i^{\text{norm}} &= \frac{I_i - \min(I)}{\max(I) - \min(I)}, \ \ \ \ \ \ \ 
    I_i^{\text{final}} &= \begin{cases} 
    I_i^{\text{norm}} & \text{if } I_i^{\text{norm}} \geq \theta \\
    0 & \text{otherwise}
    \end{cases}
\end{aligned}
\end{equation}
where $\theta$ is the importance threshold that determines which nodes are considered significant.

\subsubsection{Model Architecture}

The model consists of three stacked IP Layers with GraphNorm, Dropout, and ReLU between layers. The final prediction is computed as follows:
\begin{equation}
p = \sigma\bigg(\mathbf{W}f \big(\frac{1}{|V|} \sum{v \in V} \mathbf{h}^{(L)}_v \big) + b_f \bigg)
\end{equation}
where $p$ is the positive class probability and $\mathbf{h}^{(L)}_v$ is the final node representation.
The loss function combines binary cross entropy (BCE) and importance regularization:
\begin{equation}
\mathcal{L} = \mathcal{L}_{\text{BCE}} + w_{\text{imp}} \cdot \mathcal{L}_{\text{imp}}
\end{equation}
where $\mathcal{L}_{\text{imp}}$ is L1 regularization on importance scores.

\section{Experiments}
\label{sec:experiment}

\subsection{Dataset}

571We processed the NCI-60 dataset~\citep{shoemaker2006nci60} using rcellminer~\citep{luna2016rcellminer}, applying a threshold of -4.595 to log-transformed IC50 (50\% inhibitory concentration) values to initially achieve a balanced 50:50 drug sensitive/resistance labels ratio. After selecting drugs with NSC (National Service Center number) identifiers, the final dataset comprised 53,852 entries (36,171 positive, 17,681 negative). For zero-shot prediction, we split the data using 70\% cell lines and 60\% drugs for training/validation (571 drugs, 42 cell lines) and the rest for testing (381 drugs, 18 cell lines). This resulted in 18,067 training, 4,516 validation, and 6,525 test samples.

\subsection{Prediction Performance}

To evaluate \mname, we compared it against several baseline methods, including five traditional ML approaches, two current research methods, and 3 GNNs without an IP layer. Table~\ref{tab:performance_comparison} presents the performance metrics for each method, averaged over five independent runs.

\begin{table}[h]
\centering
\resizebox{\textwidth}{!}{
\begin{tabular}{c||c|c|ccccc}
\toprule
& \textbf{Methods} & Explainability & \textbf{ROC-AUC} ($\uparrow$) & \textbf{PR-AUC} ($\uparrow$) & \textbf{Accuracy} ($\uparrow$) & \textbf{Precision}
($\uparrow$) & \textbf{Specificity} ($\uparrow$) \\
\midrule
\multirow{7}{*}{\rotatebox[origin=c]{90}{\textbf{Baseline}}} & RF & \stackunder{Feature}{Importance} & \stackunder{0.788}{\scriptsize (±0.001)} & \stackunder{0.892}{\scriptsize (±0.002)} & \stackunder{0.716}{\scriptsize (±0.002)} & \stackunder{0.726}{\scriptsize (±0.002)} & \stackunder{{0.632}}{\scriptsize (±0.003)} \\
& LightGBM & \stackunder{Feature}{Importance} & \stackunder{0.790}{\scriptsize (±0.000)} & \stackunder{0.870}{\scriptsize (±0.000)} & \stackunder{0.747}{\scriptsize (±0.000)} & \stackunder{0.769}{\scriptsize (±0.000)} & \stackunder{0.457}{\scriptsize (±0.000)} \\
& MLP & - & \stackunder{0.750}{\scriptsize (±0.010)} & \stackunder{0.838}{\scriptsize (±0.006)} & \stackunder{0.710}{\scriptsize (±0.004)} & \stackunder{0.721}{\scriptsize (±0.009)} & \stackunder{0.271}{\scriptsize (±0.051)} \\
& MPNN & - & \stackunder{0.792}{\scriptsize (±0.013)} & \stackunder{0.892}{\scriptsize (±0.006)} & \stackunder{0.728}{\scriptsize (±0.009)} & \stackunder{0.726}{\scriptsize (±0.011)} & \stackunder{0.571}{\scriptsize (±0.044)} \\
& GCN & - & \stackunder{0.766}{\scriptsize (±0.019)} & \stackunder{0.872}{\scriptsize (±0.010)} & \stackunder{0.710}{\scriptsize (±0.021)} & \stackunder{0.710}{\scriptsize (±0.020)} & \stackunder{0.559}{\scriptsize (±0.029)} \\ \midrule

\multirow{3}{*}{\rotatebox[origin=c]{90}{\stackunder{{\small\textbf{Previous}}}{\small\textbf{Research}}}} & DeepDSC & - & \stackunder{0.713}{\scriptsize (±0.014)} & \stackunder{0.783}{\scriptsize (±0.009)} & \stackunder{\textbf{0.751}}{\scriptsize (±0.011)} & \stackunder{\textbf{0.807}}{\scriptsize (±0.009)} & \stackunder{0.599}{\scriptsize (±0.021)} \\
& MOFGCN & - & \stackunder{0.492}{\scriptsize (±0.000)} & \stackunder{0.666}{\scriptsize (±0.000)} & \stackunder{0.355}{\scriptsize (±0.000)} & \stackunder{0.650}{\scriptsize (±0.000)} & \stackunder{\textbf{0.901}}{\scriptsize (±0.000)} \\ \midrule

\multirow{3}{*}{\rotatebox[origin=c]{90}{\stackunder{\textbf{Ablation}}{\textbf{w/o IP layer}}}}
& GAT & - & \stackunder{0.758}{\scriptsize (±0.019)} & \stackunder{0.868}{\scriptsize (±0.013)} & \stackunder{0.703}{\scriptsize (±0.007)} & \stackunder{0.687}{\scriptsize (±0.011)} & \stackunder{0.395}{\scriptsize (±0.066)} \\
& GT & - & \stackunder{{0.774}}{\scriptsize (±0.019)} & \stackunder{{0.874}}{\scriptsize (±0.016)} & \stackunder{0.717}{\scriptsize (±0.020)} & \stackunder{0.717}{\scriptsize (±0.019)} & \stackunder{0.566}{\scriptsize (±0.034)} \\
& GINE & - & \stackunder{0.750}{\scriptsize (±0.019)} & \stackunder{0.794}{\scriptsize (±0.014)} & \stackunder{0.700}{\scriptsize (±0.018)} & \stackunder{0.670}{\scriptsize (±0.018)} & \stackunder{0.336}{\scriptsize (±0.037)} \\

\midrule
\multirow{3}{*}{\rotatebox[origin=c]{90}{\textbf{\mname}}}
& GAT & Node Importance & \stackunder{0.789}{\scriptsize (±0.006)} & \stackunder{0.892}{\scriptsize (±0.005)} & \stackunder{0.720}{\scriptsize (±0.012)} & \stackunder{0.717}{\scriptsize (±0.013)} & \stackunder{0.547}{\scriptsize (±0.048)} \\
& GT & Node Importance & \stackunder{\textbf{0.796}}{\scriptsize (±0.006)} & \stackunder{\textbf{0.894}}{\scriptsize (±0.001)} & \stackunder{0.724}{\scriptsize (±0.005)} & \stackunder{0.719}{\scriptsize (±0.006)} & \stackunder{0.548}{\scriptsize (±0.019)} \\
& GINE & Node Importance & \stackunder{0.790}{\scriptsize (±0.003)} & \stackunder{0.891}{\scriptsize (±0.001)} & \stackunder{0.730}{\scriptsize (±0.012)} & \stackunder{0.728}{\scriptsize (±0.015)} & \stackunder{0.575}{\scriptsize (±0.056)} \\
\bottomrule
\end{tabular}
}

\caption{\textbf{Predictive Performance Comparison for Binary Classification.} Results show averages of 5 independent runs with standard deviations in parentheses. The best values for each metric are in \textbf{bold}. Abbreviations: ROC-AUC: Receiver Operating Characteristic Area Under the Curve, PR-AUC: Precision-Recall Area Under the Curve, RF: Random Forest, MLP: Multiple Layer Perceptron, MPNN: Message-Passing Neural Network, GCN: Graph Convolutional Networks, MOFGCN: Multi-Omics Data Fusion and Graph Convolution Network, GAT: Graph Attention Network, GT: Graph Transformer, GINE: Graph Isomorphism Network with Edge features. Feature Importance: A measure of how much each feature contributes to a model's predictions.}
\label{tab:performance_comparison}
  \vspace{-10pt}
\end{table}

\begin{table}[ht]
\centering
\small
\begin{tabular}{c||c||cc}
\hline
& \textbf{Methods} & \textbf{ROC-AUC} & \textbf{PR-AUC} \\
\hline
\multirow{6}{*}{\rotatebox[origin=c]{90}{\textbf{Baseline}}} & RF & 0.788 & 0.892 \\
& LightGBM & 0.790 & 0.870 \\
&MPNN & 0.792 & 0.892 \\
&GCN & 0.766 & 0.872 \\
&DeepDSC & 0.713 & 0.783 \\ 
&MOFGCN & 0.492 & 0.666 \\ \hline
\multirow{3}{*}{\rotatebox[origin=c]{90}{\textbf{w/o IP}}} & GAT & 0.758 & 0.868 \\
 & Graph Transformer (GT) & 0.774 & 0.874 \\
& GINE & 0.750 & 0.794 \\ \hline
\multirow{3}{*}{\rotatebox[origin=c]{90}{\textbf{Ours}}} & GraphPINE (GAT) & 0.789 & 0.892 \\
&\textbf{GraphPINE (GT)} & \textbf{0.796} & \textbf{0.894} \\
& GraphPINE (GINE) & 0.790 & 0.891 \\
\hline
\end{tabular}
\caption{Predictive performance comparison using ROC-AUC and PR-AUC metrics. The best results among the Graph Transformer (GT) models are highlighted in bold. ``w/o IP'' denotes ablation studies conducted without IP layers.} 
\label{tab:performance}
\end{table}

Our \mname~model, particularly the GT variant, demonstrates superior performance across multiple metrics. Given the imbalanced nature of our dataset, we place particular emphasis on the PR-AUC and ROC-AUC scores as the most critical evaluation metrics. Notably, \mname~(GT) achieves the highest PR-AUC (0.894) and ROC-AUC (0.796), underscoring its effectiveness in handling imbalanced data. While DeepDSC shows higher accuracy (0.751) and precision (0.807), \mname~(GT)'s balanced performance across multiple metrics indicates its robust ability to effectively discriminate between classes.

MOFGCN exhibits a performance pattern with a high specificity (0.901) but poor performance across other metrics (ROC-AUC: 0.492, PR-AUC: 0.666, Accuracy: 0.355). This suggests that while the model excels at identifying resistance, it does so at the expense of overall classification performance, indicating a highly imbalanced prediction behavior that limits its utility.

The ablation study demonstrates the significant impact of the IP layer across all architectures. The GT variant achieves the best performance with PR-AUC of 0.894 and ROC-AUC of 0.796, representing improvements of 2.29\% and 2.74\% from its baseline scores of 0.874 and 0.774, respectively. The GAT architecture exhibits notable enhancements, with PR-AUC increasing by 2.74\% (from 0.868 to 0.892) and ROC-AUC by 4.04\% (from 0.758 to 0.789). Most remarkably, the GINE architecture shows the most substantial improvement, with PR-AUC increasing by 12.29\% (from 0.794 to 0.891) and ROC-AUC by 5.36\% (from 0.750 to 0.790), demonstrating the IP layer's effectiveness in enhancing model performance.

It is worth noting that all variants of \mname~(GINE, GAT, and GT) show low standard deviations across runs, indicating the stability and reliability of our proposed method. This consistency is valuable when dealing with imbalanced datasets, as it suggests that our model's performance is robust across different data splits and initializations.

\subsection{Interpretability Analysis}

\begin{figure}[h]
\centering
\begin{minipage}{0.5\textwidth}
    \centering
    \includegraphics[width=\linewidth]{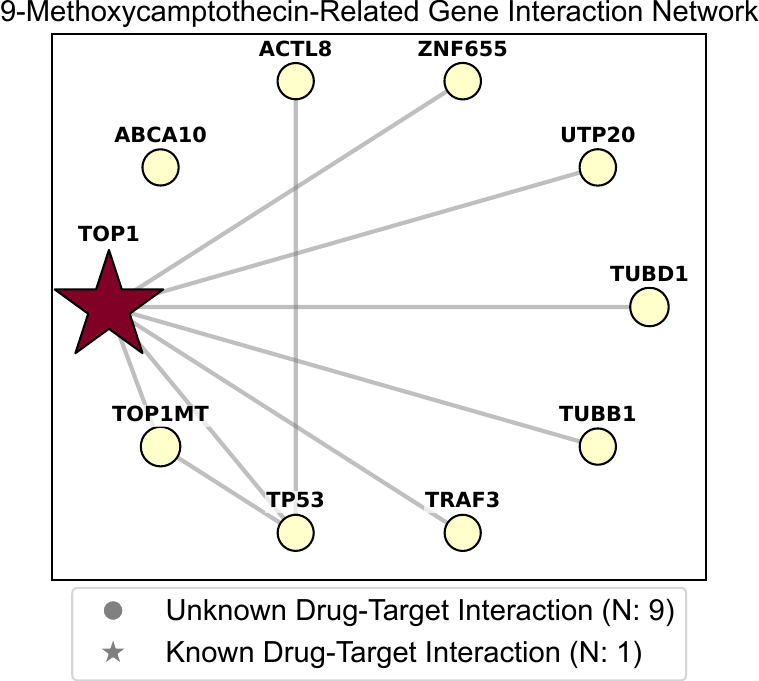} 
    \caption{
        \textbf{Gene importance scores for 9-Methoxycamptothecin.} Node size describes the propagated gene importance, and node color shows the initial DTI score.
    }
    \label{fig:interpretability}
\end{minipage}\hfill
\begin{minipage}{0.48\textwidth}
    \centering
    \resizebox{\textwidth}{!}{
    \begin{tabular}{c|c|c|c|c}
    \toprule
    Rank & \stackunder{Initial}{Importance} & Gene & PMIDs & Relationship\\
    \midrule
    1 & 1 & TOP1& 29312794... & Target\\ 
    2 & - & TOP1MT& 24890608... & Indirect\\ 
    3 & - & TUBD1& - & -\\ 
    4 & - & ZNF655& -& -\\ 
    5 & - & UTP20& -& -\\ 
    6 & - & TUBB1& -& -\\ 
    7 & - & ACTL8& -& -\\ 
    8 & - & ABCA10& 10606239 & Indirect\\ 
    9 & - & TRAF3& -& -\\ 
    10 & - & TP53& 12082016... & Indirect\\
    \bottomrule
    \end{tabular}
    }    
    \captionof{table}{
    \textbf{Top 10 predicted important genes for 9-Methoxycamptothecin and related literature.} (-) represents no initial DTI (0), and  (...) describes multiple papers.
    Target: Genes encoding proteins that directly bind to and interact with the drug. Indirect: Genes that do not encode proteins that physically interact with the drug but are involved in its mechanism of action, pathway, or response.}
    \label{tab:evidence}
  \vspace{-20pt}
\end{minipage}
\end{figure}

\mname~assigns importance scores to each gene, indicating their relative significance in predicting drug responses. Figure~\ref{fig:interpretability} illustrates the gene interaction network associated with 9-Methoxycamptothecin (MCPT), a DNA damage-related anticancer drug and derivative of camptothecin (CPT). In this network, the size of each node reflects the propagated gene importance after prediction, while the node shape differentiates between known DTIs (denoted by a star) and unknown interaction partners (denoted by a circle). The color of the nodes represents known DTI scores. The known target of 9-MCPT is TOP1; other genes may affect response directly or indirectly. Figure~\ref{fig:interpretability}  shows that the known target, TOP1, has the highest DTI score and propagated importance, and other genes have propagated importance but are low compared with TOP1. ABCA10 lacks an edge because it is not among the top interactions shown.

Table~\ref{tab:evidence} lists the top 10 important predictive genes related to 9-MCPT, including TOP1. Although TOP1MT is not known as a target of 9-MCPT, CPT and CPT derivatives can trap TOP1MT-DNA cleavage complexes~\citep{zhang2008mitochondrial}, suggesting 9-MCPT may be indirectly effective against TOP1MT. Additionally, there is an established association between CPT and ABC transporters, making it plausible that ABCA10 might also be related to 9-MCPT activity. Moreover, the efficacy of 9-MCPT may be influenced by the status of TP53, which modulates cellular responses to DNA damage~\citep{abuetabh2022dna}.

These results demonstrate that \mname~can identify biologically relevant gene relationships from gene-gene networks by incorporating prior DTI information. While TOP1MT and TP53 are established as functionally related genes but not known drug targets, our model captures several of these secondary relationships, highlighting its ability to detect both direct and indirect drug-gene associations.

\subsection{Evaluation of Importance Score Propagation}

\begin{wrapfigure}{r}{0.5\textwidth}
\centering
    \begin{tabular}{c|c}
    Metric & Value \\ \toprule
    Cosine sim. & 0.87 \\ 
     Spearman corr. & 0.82 \\ 
     Rank changes & 90.42\% \\ 
     Avg. shift & ±67.02 \\ 
     Max up & 946 \\ 
     Max down & -932 \\ \bottomrule
    \end{tabular}
    \captionof{table}{\textbf{Differences in Node (Gene) Ranks Before and After Propagation.} 
    Cosine sim.: Cosine similarity between initial/propagated importance rank. Spearman corr.: Spearman Rank correlation between initial/propagated importance rank. Rank changes: The percentage of genes whose ranks changed after propagation. Avg. shift: The average rank shift. Max up/down: Maximum upward/downward rank mobility. 
    }
    \vspace{-10pt}
    \label{tab:metrics}
\end{wrapfigure}

To understand the extent to which our importance propagation affects our initial importance scores, we analyzed 6000 randomly selected drug-cell combinations (389 unique drugs × 26 cell lines) across 5181 genes. Our prior knowledge interaction data is highly sparse; each drug was associated with between 1 and 956 interactors (an average of 39.86 interactions). Appendix~\ref{app:import}  includes a distribution of the number of interactions (Table~\ref{tab:metrics}).   Importance scores of 0 imply the absence of an interaction, and non-zero values imply an interaction. Therefore, we first examine the extent to which our propagation method increased non-zero values. We observed that non-zero values increased from 0.77\% to 39.8\% after propagation; this increased the average number of non-zero values per drug from 39.86 to 2061.81 (Appendix~\ref{app:import}). Next, we examined how much individual non-zero values were altered by propagation using a similarity comparison and a rank change analysis. For the similarity analysis (using cosine similarity and Spearman rank correlation), we observe a high but not perfect correlation (0.89 and 0.82, respectively); this suggests importance values that are updated as part of the training process. Approximately 90\% of importance values showed some rank change as an effect of propagation with an average shift of ±67.02 (maximum +946/-932), Next, we considered the situation of starting with random initial importance values, and asked if training shifts these values toward our prior knowledge-derived importance values.



\section{Discussion}

We introduced \mname, an interpretable GNN architecture featuring an ``Importance Propagation Layer''. Equations \ref{eq:gat} through \ref{eq:propagation} show how node features and importance scores are updated through training while preserving prior knowledge for stability and adaptability.

Our analysis demonstrates that GraphPINE effectively balances initial knowledge with learned patterns. While the model starts with initial node importance values, the propagation mechanism successfully discovers new relationships (increasing interactions from 0.77\% to 39.8\%) while maintaining meaningful initial characteristics (0.9 cosine similarity). This balance enables both stability from prior knowledge and adaptability to new patterns. Future work could explore additional information sources, such as protein-protein interaction networks, to further enhance this capability.

While our study focuses on DRP, the \mname~framework holds potential for a wide range of applications in fields that involve complex network structures with inherent node importance. For instance, PageRank scores could be used as initial importance values to enhance the propagation of search relevance among web pages in graph analysis. 

\section{Funding}
This research was supported in part by the Division of Intramural Research (DIR) of the National Library of Medicine (NLM, ZIALM240126), National Institutes of Health (NIH) (ZIALM240126).

\clearpage

\bibliography{iclr2025_conference}
\bibliographystyle{iclr2025_conference}

\clearpage

\appendix

\section{Implementation Details and Hyperparameter Tuning}\label{app:imp_and_tuning}

\subsection{Data Preprocessing and Network Construction}\label{app:data}

We integrated multiple data sources to create a comprehensive gene-gene interaction network and DTI dataset. Our approach involves several key steps.

\subsubsection{Data Integration}\label{app:multi}
Let $G = {g_1, g_2, ..., g_n}$ be the set of all genes, and $D = {d_1, d_2, ..., d_m}$ be the set of all drugs. We collected data from various sources. From NCI-60 cell lines, we obtained multi-omics data including gene expression ($X_{exp} \in \mathbb{R}^{n \times c}$), methylation ($X_{met} \in \mathbb{R}^{n \times c}$), mutation ($X_{mut} \in {0, 1}^{n \times c}$), and copy number variation (CNV) ($X_{cnv} \in \mathbb{R}^{n \times c}$), where $n$ is the number of genes and $c$ is the number of cell lines. Gene-gene interaction data ($E_{gg} \subseteq G \times G$) was sourced from PathwayCommons~\citep{cerami2010pathway, rodchenkov2020pathway}, containing various types of interactions such as catalysis-precedes, controls-expression-of, controls-phosphorylation-of, controls-state-change-of, controls-transport-of, in-complex-with, and interacts-with. DTI data ($E_{dti} \subseteq D \times G$) was collected from multiple sources, including the CTD, DrugBank, DGIdb, STITCH, and the KIBA dataset.

\subsubsection{Gene-Gene Network Construction}\label{app:genegene}
We selected a subset of genes $G' \subseteq G$ based on three criteria. (1) First, we considered variance in multi-omics data. For each data source $s \in \{\text{exp}, \text{met}, \text{mut}, \text{cnv}\}$ where exp represents gene expression, met represents methylation, mut represents mutation, and cnv represents copy number variation,  we computed the variance for each gene across cell lines:
\begin{equation}
\text{var}_s(g_i) = \frac{1}{c-1} \sum_{j=1}^c (X_{s_{ij}} - \bar{X}{s_i})^2
\end{equation}
We selected the top 3000 genes with the highest variance for each data source. (2) Second, we computed network centrality, calculating the degree of centrality for each gene in the initial interaction network:
\begin{equation}
\text{centrality}(g_i) = \frac{|{(g_i, g_j) \in E_{gg} \vee (g_j, g_i) \in E_{gg}}|}{|G| - 1}
\end{equation}
We selected the top 3000 genes with the highest centrality. (3) Third, we considered DTI frequency, calculating the frequency of each gene in the DTI data:
\begin{equation}
\text{freq}_{\text{dti}}(g_i) = |{(d_j, g_i) \in E_{dti}}|
\end{equation}
We selected the top 3000 genes with the highest DTI frequency. The final set of genes $G'$ was the union of these selections, resulting in 5,181 genes. We then constructed the gene-gene interaction network $G' = (V', E')$, where $V' = G'$ and $E' = E_{gg} \cap (G' \times G')$, containing 630,632 interactions.

\subsubsection{Edge Encoding}\label{app:edge}

Each interaction between genes is categorized into one of seven types based on the information from PathwayCommons:``catalysis-precedes'', ``controls-expression-of'', ``controls-phosphorylation-of'', ``controls-state-change-of'', ``controls-transport-of'', ``in-complex-with'', and ``interacts-with''. These interaction types were encoded as one-hot vectors.

Let $T = \{t_1, t_2, ..., t_7\}$ represent the set of all interaction types. For each edge $e \in E'$, a binary vector $v_e \in \{0, 1\}^7$ was created, where each element corresponds to a specific interaction type:
\begin{equation}
    v_e[i] = \begin{cases}
        1 & \text{if edge $e$ has interaction type $t_i$} \\
        0 & \text{otherwise}. 
    \end{cases}
\end{equation}

\subsubsection{Multi-omics Data Preprocessing}
We focused on normalizing gene expression data through several steps. First, we converted the data to Transcripts Per Million (TPM):
\begin{equation}
\text{TPM}_{ij} = \frac{X_{\text{exp}_{ij}}}{\sum_{i=1}^n X_{\text{exp}_{ij}}} \times 10^6. 
\end{equation}
Next, we applied a Log2 transformation:
\begin{equation}
X'_{\text{exp}{ij}} = \log_2(\text{TPM}_{ij} + 1). 
\end{equation}
Finally, we performed Winsorization. Let $q_{0.1}$ and $q_{99.9}$ be the 0.1 and 99.9 percentiles of $X'{exp}$. We applied:
\begin{equation}
        X''_{\text{exp}_{ij}} = \begin{cases}
            q_{0.1} & \text{if } X'_{\text{exp}_{ij}} < q_{0.1} \\
            q_{99.9} & \text{if } X'_{\text{exp}_{ij}} > q_{99.9} \\
            X'_{\text{exp}_{ij}} & \text{otherwise}. 
\end{cases}
\end{equation}

These steps ensured our gene expression data was normalized and scaled for further analysis.
We then created 4-dimensional feature vectors for each gene in each cell line:

\begin{equation}
    X_i = [X''_{\text{exp}_i}, X_{\text{met}_i}, X_{\text{mut}_i}, X_{\text{cnv}_i}]. 
\end{equation}

\subsection{Implementation Details}\label{app:implementation}
The \mname~model was implemented using Python 3.10, PyTorch 2.4.0, and PyTorch Geometric 2.5.3, leveraging their efficient deep learning and graph processing capabilities. We employed the Adam optimizer for training, with a learning rate of 0.001 and a batch size of 32. The model architecture incorporates 3 Importance Propagation Layers ($L=3$), each containing 64 hidden units. To balance model performance and interpretability, we set the importance regularization coefficient $\lambda$ to 0.01 and the importance threshold $\tau$ to 0.1.

All experiments were conducted on NVIDIA Tesla A100 GPUs with 80 GB memory. The average training time for \mname~was 0.2 seconds, with an inference time of 0.1 seconds per drug-cell line pair, demonstrating its feasibility for large-scale DRP tasks. To ensure reproducibility and facilitate further research, we have made our code and datasets publicly available at \url{\repo}.

\subsection{Training Procedure}\label{app:training}

The training procedure for the \mname~model is designed to optimize performance while preventing overfitting. Algorithm~\ref{alg:training_procedure} presents a detailed overview of this process.
Concretely, the \mname~training procedure involves initializing model parameters, iterating through epochs, performing forward and backward passes, computing losses, and updating parameters. The procedure also includes an early stopping mechanism to prevent overfitting.

We employ the Adam optimizer with an initial learning rate of $\eta = 10^{-3}$.

\begin{algorithm}
\caption{\mname~Training Procedure \label{alg:training_procedure}}
\begin{algorithmic}[1]
\State Initialize model parameters $\theta$
\State Initialize optimizer with learning rate $\eta$
\State Set early stopping patience $p$ and minimum delta $\delta$
\For{epoch $= 1$ to $T_{\text{total}}$}
    \For{batch in training data}
        \State Forward pass: $\hat{y}, I' = f_{\theta}(X, E, I)$
        \State Compute loss: $L = w_{\text{BCE}} \cdot \mathcal{L}_{\text{BCE}}(\hat{y}, y) + w_{\text{imp}} \cdot \mathcal{L}_{\text{imp}}(I', I)$
        \State Backward pass: Compute $\nabla_{\theta} \mathcal{L}$
        \State Update parameter using Adam optimizer. 
    \EndFor
    \State Evaluate on the validation set
    \If{validation loss improved by at least $\delta$}
        \State Reset patience counter
        \State Save the best model
    \Else
        \State Decrement patience counter
        \If{patience counter $= 0$}
            \State Early stop and return best model
        \EndIf
    \EndIf
\EndFor
\end{algorithmic}
\end{algorithm}

\subsection{Hyperparameter Tuning}\label{app:tuning}
To optimize the performance of our \mname~model, we conducted extensive hyperparameter tuning using Optuna~\citep{Akiba2019Optuna}, an efficient hyperparameter optimization framework. We utilized MLflow for experiment tracking and logging, ensuring comprehensive documentation of our optimization process.

Our hyperparameter search space encompassed key model parameters, including the number of epochs (1-3), number of attention heads (1, 2, 4), number of GNN layers (2-4), dropout rate (0.1-0.3), importance decay (0.7-0.9), importance threshold (1e-5 to 1e-3), hidden channel size (16, 32), BCE weight (0.9-1.1), importance regularization weight (0.005-0.02), and learning rate (0.001-0.1). The batch size was initially set to 5, with a dynamic reduction mechanism implemented to handle potential memory constraints.

The optimization process consisted of 20 trials, each involving the following steps: (1) hyperparameter suggestion by Optuna, (2) \mname~model initialization with the suggested configuration, (3) model training and validation, and (4) reporting of the minimum validation loss as the objective value for optimization. This systematic approach allowed us to identify the optimal hyperparameter configuration that balanced model performance and computational efficiency.

Throughout the implementation and tuning process, we leveraged several key libraries and tools. PyTorch served as the foundation for building and training our neural network model. Optuna facilitated efficient hyperparameter optimization, while MLflow provided robust experiment tracking and logging capabilities. We also utilized NumPy for numerical computations and Pandas for data manipulation and analysis, ensuring a comprehensive and efficient development environment.

This rigorous implementation and tuning process enabled us to develop a highly optimized \mname~model capable of accurate and interpretable DRPs. The combination of advanced deep learning techniques, efficient hyperparameter optimization, and careful implementation considerations resulted in a model that balances performance, interpretability, and computational efficiency.

\subsection{Baseline Setting} \label{sec:baseline_setting}

We implemented three baseline models for comparison: Random Forest (RF), LightGBM, and Multiple Layer Perceptron (MLP). All models were trained on the same dataset, which combined gene expression, methylation, mutation, copy number variation, and drug-target interaction data.

\paragraph{Random Forest (RF):} \label{para:rf}
We used \texttt{scikit-learn}'s \texttt{RandomForestClassifier} with hyperparameters optimized via \texttt{Optuna}. The key hyperparameters included the number of estimators (100--1000), max depth (10--100), min samples split (2--20), min samples leaf (1--10), and max features (None, \texttt{"sqrt"}, or \texttt{"log2"}).

\paragraph{LightGBM:} \label{para:lightgbm}
We implemented LightGBM~\citep{ke2017lightgbm} with binary classification objective and log loss metric. Hyperparameters were tuned using \texttt{Optuna}, including num\_leaves (31--255), learning\_rate (1e-3 to 1.0), feature\_fraction (0.1--1.0), bagging\_fraction (0.1--1.0), bagging\_freq (1--7), min\_child\_samples (5--100), lambda\_l1 and lambda\_l2 (1e-8 to 10.0), and num\_boost\_round (100--2000).

\paragraph{Multiple Layer Perceptron (MLP):} \label{para:mlp}

We created a \texttt{PyTorch}-based MLP with a flexible architecture. Hyperparameters optimized via \texttt{Optuna} included the number of layers (2--5), hidden dimensions (64--512 units per layer), learning rate (1e-5 to 1e-1), batch size (32, 64, 128, or 256), dropout rate (0.1--0.5), and normalization type (batch or layer normalization).

\paragraph{DeepDSC and MOFGCN:}

For DeepDSC, we follow the original architecture consisting of a stacked autoencoder followed by a feed-forward network. The encoder comprises three hidden layers (2,000, 1,000, and 500 units), while the decoder mirrors this with hidden layers of 1,000 and 2,000 units. The activation function is selu for hidden layers and sigmoid for the output layer. Training employs AdaMax optimizer with a learning rate of 0.0001, gradient clipping at 1.0, and Xavier uniform initialization1.

For MOFGCN, we utilize the following hyperparameters: scale parameter $\varepsilon=2$, proximity parameter $N=11$, number of iterations $t=3$, embedding dimension $h=192$, correlation information dimension $k=36$, scaling parameter $\alpha=5.74$, learning rate $5\times10^{-4}$, and 1000 training epochs. The model uses the PyTorch framework with Adam optimizer.

Both models employ early stopping to prevent overfitting - DeepDSC with the patience of 30 epochs and MOFGCN monitoring the validation loss.

\paragraph{MPNN, GCN, and GINE:} \label{para:mpnn_gcn}
For the MPNN (Message-Passing Neural Network)~\citep{gilmer2017neural}, GCN (Graph Convolutional Network)~\citep{kipf2016semi}, and GINE (Graph Isomorphism Network with Edge features)~\citep{hu2019strategies, xu2018powerful}, we tuned the hyperparameters using the following configuration. The number of epochs (num\_epochs) was selected from \{10, 50, 100\}. The batch size was chosen from \{2, 3, 4\}. The number of GNN layers was selected from \{1, 2, 3\}. The dropout rate was selected from \{0.1, 0.2, 0.3\}. The importance decay was chosen from \{0.7, 0.8, 0.9\}. The importance threshold was selected from \{1e-5, 1e-4, 1e-3\}. The hidden channel size was selected from \{16, 32\}. The weight for the mean squared error loss was selected from \{0.9, 1.0, 1.1\}. The weight for importance regularization was selected from \{0.005, 0.01, 0.02\}. The learning rate was selected from \{0.001, 0.01, 0.1\}.

\paragraph{GAT and Graph Transformer:} \label{para:gat_graph_transformer}
For the GAT (Graph Attention Network)~\citep{velivckovic2017graph} and Graph Transformer models~\citep{yun2019graph}, we used a similar hyperparameter tuning configuration as for MPNN, GCN, and GINE. However, for GAT and Graph Transformer, we also included the number of attention heads, which was selected from \{1, 2, 4\}. This additional parameter helps in controlling the number of attention mechanisms in the model, enabling it to learn more complex representations.

For all models, we used \texttt{Optuna} for hyperparameter optimization, maximizing accuracy on the validation set. Each model was then trained five times with the best hyperparameters, and we reported the mean and standard deviation of accuracy, precision, recall, and F1 score on the test set.

The data preprocessing steps were consistent across all models, including normalization of gene expression data and concatenation of multi-omics features. This ensured a fair comparison between the baseline models and our proposed \mname~method.

\section{Experiments}\label{app:experiments}

\subsection{Dataset and Preprocessing}

In this study, we utilized a comprehensive drug response dataset containing information on multiple cell lines and compounds. The dataset was preprocessed and split to ensure a rigorous evaluation of the model's generalization capabilities. Initially, the dataset contained IC50 data for unique cell lines and unique NSC (Cancer Chemotherapy National Service Center number) identifiers for compounds.

To adapt this data for binary classification, we applied an empirically determined threshold, which was set to achieve an approximately 50:50 ratio of response to non-response using the formula below. 
\begin{equation}
\text{binarize}(x)= \begin{cases}1 & \text { if } x<\text { threshold } \\ 0 & \text { otherwise, }\end{cases},
\end{equation}
where the threshold is the hyperparameter, and we set -4.595.

This process resulted in a dataset of 331,558 entries. We then refined our dataset to focus on the 60 cell lines present in the NCI60 panel, reducing the data to 315,778 entries. Further narrowing our scope to include only the drugs used in the NCI60 project, we arrived at a final dataset of 53,852 entries.

To set up a zero-shot prediction scenario, we randomly selected 70\% of unique cell lines and 60\% of unique NSC identifiers for the training and validation sets. The remaining cell lines and NSC identifiers were used for the test set, ensuring no overlap of cell lines or compounds between the train/validation and test sets. This approach allows us to evaluate the model's ability to generalize to entirely new cell-compound combinations.

The data was split as follows: The training set comprises 18,067 entries, consisting of 571 unique drugs (NSCs) and 42 unique cell lines. The validation set contains 4,516 entries, utilizing the same 571 drugs and 42 cell lines as the training set. The test set includes 6,525 entries, encompassing 381 unique drugs and 18 unique cell lines.

Notably, while the training and validation sets share common cell lines and drugs, the test set introduces new drug-cell line combinations. This configuration allows for a rigorous assessment of our model's generalization capability, enabling us to evaluate its predictive performance on unseen drug-cell line pairs.

\subsection{Evaluation Metrics}

We evaluated \mname~using a comprehensive set of metrics to assess its classification performance. The Accuracy was used to measure the overall correctness of the model's predictions across all classes. To provide a more nuanced assessment of the model's discriminative ability, we calculated the Area Under the Receiver Operating Characteristic curve (ROC-AUC) and the Area Under the Precision-Recall curve (PR-AUC). ROC-AUC quantifies the model's ability to distinguish between classes across various threshold settings, while PR-AUC is particularly useful for evaluating performance on imbalanced datasets.
To further characterize the model's performance on negative instances, we computed the Specificity, which measures the proportion of actual negatives correctly identified. Additionally, we calculated the Negative Predictive Value (NPV), which quantifies the proportion of negative predictions that were correct. These metrics collectively offer a thorough evaluation of \mname's ability to correctly classify both positive and negative instances, providing insights into its performance across different aspects of the classification task.

\subsection{Interpretability Analysis}\label{app:interpret}

\begin{figure}[!t]
\centering
\begin{minipage}{0.5\textwidth}
    \centering
    \includegraphics[width=\linewidth]{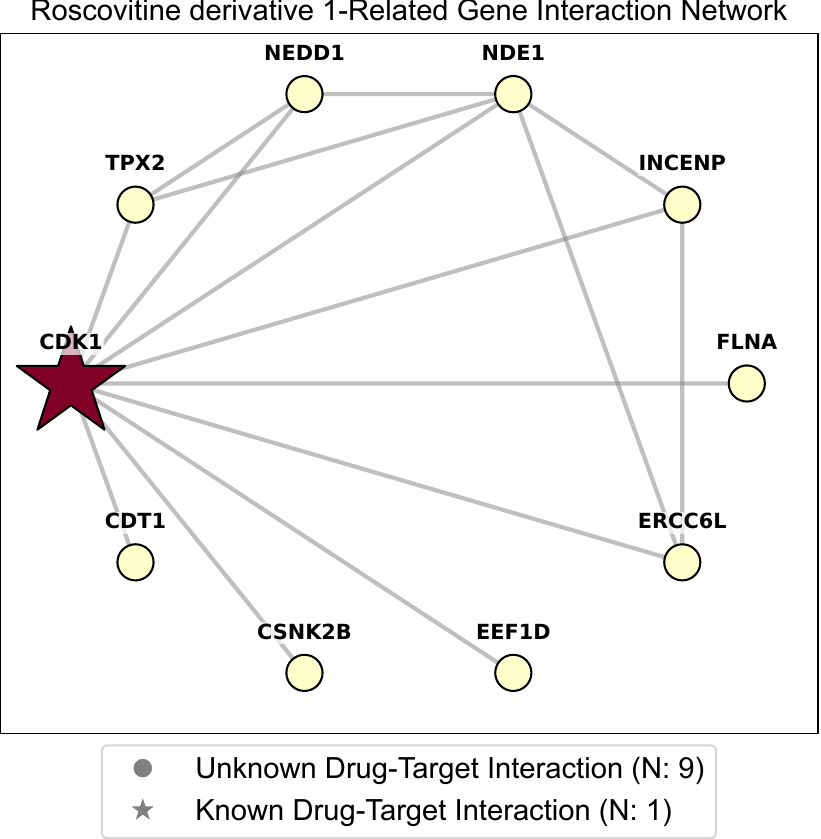} 
    \caption{Gene importance scores and interactions for Roscovitine derivative 1. Node size describes the propagated gene importance.}
    \label{fig:Roscovitine}
\end{minipage}\hfill
\begin{minipage}{0.48\textwidth}
    \centering
    \captionof{table}{Top 10 predicted important genes for Roscovitine derivative 1.}
    \resizebox{\textwidth}{!}{
    \begin{tabular}{c|c|c}
    \toprule
    Rank & Gene Name & Evidence (PMID)\\
    \midrule
1  &   CDK1 &     37635245 \\
2  &   NDE1 &     - \\
3  & INCENP &     - \\
4  &  EEF1D &     - \\
5  &  NEDD1 &     - \\
6  &   CDT1 &     35931300 \\
7  & CSNK2B &     - \\
8  &   TPX2 &     - \\
9  & ERCC6L &     - \\
10 &   FLNA &     - \\
    \bottomrule
    \end{tabular}
    }
    \label{tab:Roscovitine}
\end{minipage}
\end{figure}

Figure \ref{fig:Roscovitine} shows the predicted interaction network for a roscovitine derivative. The network contains mostly unknown interactions (9) with only one known interaction. CDK1 is highlighted as the most important predicted target gene. This suggests the roscovitine derivative may have new mechanisms of action beyond the known CDK inhibition, but CDK1 remains a key target.

Table \ref{tab:Roscovitine} lists the top 10 predicted important genes for the roscovitine derivative. CDK1 is ranked first, consistent with roscovitine's known mechanism as a CDK inhibitor. However, most other predicted genes, like NDE1, INCENP, EEF1D, etc. are new interactions without existing evidence. This suggests potential new pathways the derivative may affect beyond CDK inhibition.

\subsection{EVALUATION OF IMPORTANCE SCORE PROPAGATION}\label{app:import}

To validate our important propagation mechanism's effectiveness, we analyzed rank comparisons before/after propagation across 6000 randomly selected drug-cell combinations (389 unique drugs, 26 unique cell lines) and 5181 genes.

The initial importance density was 0.77\% with an average of 39.86 interactions per drug-cell combination. After propagation, the interaction density increased to 39.8\% (+38.96\%) with 2061.81 average interactions.

The metrics comparison revealed a high cosine similarity of 0.9, indicating that 90\% of genes maintained their original characteristics post-propagation. While the overall Spearman rank correlation was low due to zero entries, non-zero entries showed a strong correlation of 0.81, confirming the preservation of meaningful relationships during network expansion.

99.98\% of genes showed rank changes, with an average shift of ±1156.82 positions (maximum: +2658, minimum: -2590). This substantial change, combined with high similarity (0.90) to the original data, indicates the successful discovery of hidden connections while maintaining data integrity.

For non-zero DTI entries specifically, 90.5\% of genes changed ranks with an average shift of ±69.71 (maximum: +946, minimum: -932). This demonstrates that our model modifies rankings for both zero and non-zero entries while preserving cosine similarity and rank correlation.

\begin{wrapfigure}{r}{0.5\textwidth}
    \centering
    \includegraphics[width=\linewidth]{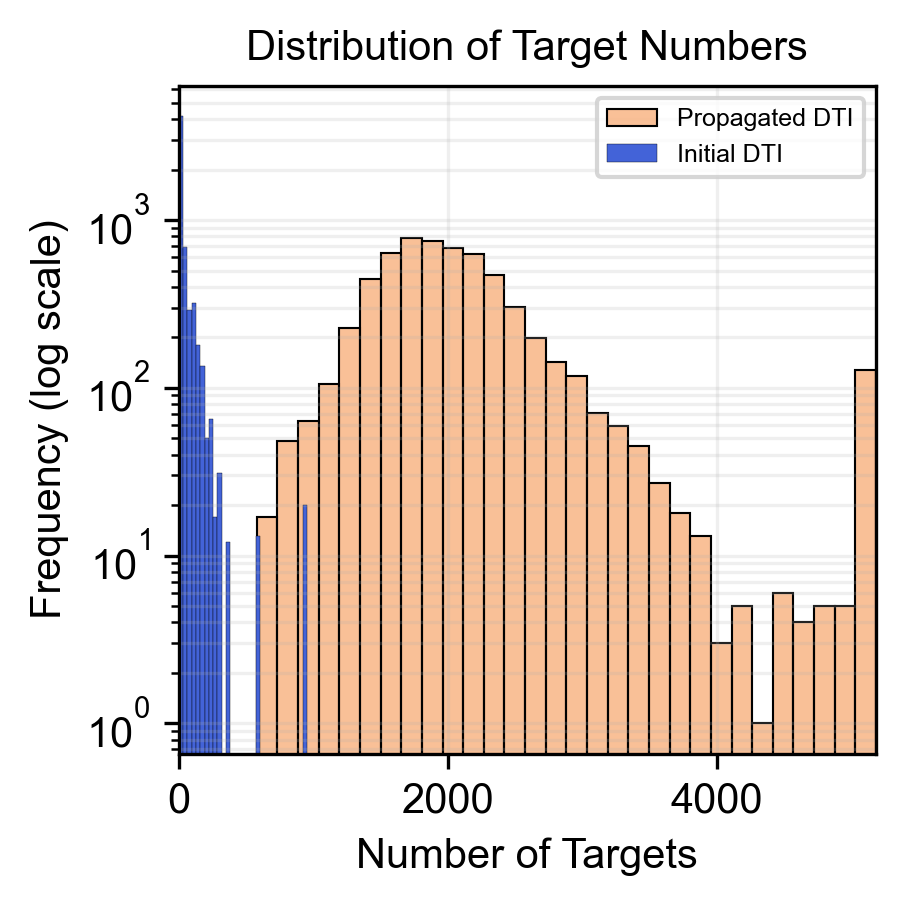} 
    \caption{
Distribution of Interactions Numbers Before/After Propagation. Initial interactions (blue) show a concentrated distribution near zero interactions, while Propagated interactions (orange) demonstrate a broader distribution centered around 2000 interactions.
} 
\vspace{-5mm}
\end{wrapfigure}

\end{document}